Chapter 11

# Conversion of Lexicon-Grammar tables to LMF: application to French[1]

## 11.1. Motivation

In this chapter, we describe the first experiment of conversion of Lexicon-Grammar (LG) tables for French verbs into the Lexical Markup Framework (LMF) format. The LG of the French language is currently one of the major sources of lexical and syntactic information for French. Its conversion into an interoperable representation format according to the LMF standard makes it usable in different contexts, thus contributing to the standardization and interoperability of natural language processing (NLP) dictionaries. We briefly introduce the LG and the derived dictionaries; we analyse the main difficulties faced during the conversion; and we describe the resulting resource.

## 11.2. The Lexicon-Grammar

### 11.2.1. *Lexicon-Grammar tables*

The LG takes the form of tables dedicated to French and to other languages, such as Italian, Portuguese, Modern Greek and Korean. Its development was initiated as early as the 1970s by Maurice Gross, at the Laboratoire d'automatique documentaire et linguistique (LADL) [GRO 75; BOO 76; GUI 92].

---

[1] Chapter written by Éric Laporte, Elsa Tolone and Matthieu Constant.

The theoretical principles underlying the LG are inspired by [HAR 57]: description focuses on directly observable surface; theoretical notions and hypotheses are used with parsimony; the resulting description has sometimes been used as a repository of theory-neutral information [HAT 98].

The LG prioritizes the readability of the dictionary for human construction and updating by linguists. Lexical information is represented in tables describing classes. Each class puts together elements of a given part of speech or lexico-grammatical category (for a given language) that share a certain number of defining features, which usually concern subcategorization information. Corresponding tables are represented as matrices: each row corresponds to a lexical item of the class; each column lists a feature that may be valid or not for the different members of the class; at the intersection of a row and a column, the + (respectively −) symbol indicates that the feature corresponding to the column is valid (respectively not valid) for the lexical entry corresponding to the row. Features are represented by mnemonic identifiers. This compact format is dedicated to manual construction and updating.

As far as the French language is concerned, the construction of the LG is coordinated by Université Paris-Est [LEC 02]. A total of 67 tables for simple verbs (13 900 lexical items) have been developed, as well as 81 tables for predicative nouns,[2] 69 tables for (mostly verbal and adjectival) idioms and 32 tables for (simple and idiomatic) adverbs. However, the experiment reported here was limited to verbs. All tables are fully available[3] under a free license (LGPL-LR). Figure 11.1 shows a sample of a verb class from [BOO 76].

| N0 =: Nhum | N0 =: N-hum | N0 =: Nnr | N0 =: V-inf W | N0 être V-n | Ppv | Ppv =: se figé | Ppv =: en figé | Ppv =: y figé | Nég | <ENT> | N0 être V-ant | N0 est Vpp | N0 V de N0pc | [extrap] | Nactif V N0 | <OPT> |
|---|---|---|---|---|---|---|---|---|---|---|---|---|---|---|---|---|
| ~ | ~ | ~ | ~ | ~ | <E> | - | - | - | - | barboter | ~ | ~ | ~ | ~ | ~ | Le gaz barbote dans l'eau |
| + | + | - | - | - | <E> | - | - | - | - | barboter | - | - | - | + | ~ | Max barbote dans l'eau |
| ~ | ~ | ~ | ~ | ~ | <E> | - | - | - | - | basculer | ~ | ~ | ~ | ~ | ~ | La chaise bascule |
| - | + | - | - | - | <E> | - | - | - | - | battre | - | - | + | - | ~ | Son coeur bat |
| - | + | - | - | - | <E> | - | - | - | - | béer | + | - | - | + | ~ | Sa bouche bée |
| - | + | - | - | - | <E> | - | - | - | - | blouser | + | - | - | - | ~ | Le chemisier blouse |
| ~ | ~ | ~ | ~ | ~ | <E> | - | - | - | - | boiter | ~ | ~ | ~ | ~ | ~ | Cette chaise boite |
| ~ | ~ | ~ | ~ | ~ | <E> | - | - | - | - | bomber | ~ | ~ | ~ | ~ | ~ | La voiture bombe |
| ~ | ~ | ~ | ~ | ~ | <E> | - | - | - | - | boucler | ~ | ~ | ~ | ~ | ~ | Le programme boucle |
| - | + | - | - | - | <E> | - | - | - | - | bouffer | + | - | - | - | ~ | Ses manches bouffent |
| ~ | ~ | ~ | ~ | ~ | <E> | - | - | - | - | bouger | ~ | ~ | ~ | ~ | ~ | La dent bouge |
| ~ | ~ | ~ | ~ | ~ | <E> | - | - | - | - | bouillir | ~ | ~ | ~ | ~ | ~ | L'eau bout à cent degrés |

**Figure 11.1.** *A sample of the table of verb class 31R.*

[2] A predicative noun is a noun that acts as the predicate in a predicate/arguments structure.

[3] http://infolingu.univ-mlv.fr/english (Language Resources > Lexicon-Grammar > Download).

Recent work made the syntactic features of the LG consistent and explicit. For each category, a ‘table of classes’ inventories the syntactic features and classes defined for this category [TOL 11a]. At the intersection of a row and a column, the + (respectively −) symbol indicates that the corresponding feature is valid (respectively not valid) for all the items in the class. The ‘o’ symbol indicates that the feature is explicitly coded in the corresponding table, because it is valid only for some of its entries. The ‘O’ symbol means that the feature should be encoded for the same reason, but is not yet listed in the table. Finally, the ‘?’ symbol means that the cell has not been filled in yet.

### 11.2.2. *The LGLex dictionary*

Thanks to this work, it was possible to derive a structured version of the LG tables: the *LGLex* dictionary, available in text or Extended Markup Language (XML) format [CON 10]. The *LGlex* format is structured on the notion of syntactic feature, but closer to current standards in NLP: features are organized into a tree; negative information is not represented. *LGLex* is computed by the *LGExtract* tool from a set of LG tables of a given category, the corresponding table of classes, and a configuration file that provides information on each feature. Thus, new versions of *LGLex* can be generated when tables are updated. Both the dictionary and the extractor are fully available[4] under a free license (LGPL-LR).

### 11.2.3. *The LGLex-Lefff dictionary*

As opposed to the LG, the Alexina format, that is that of the Le*fff* syntactic lexicon [SAG 10], is based on the notion of syntactic construction. The *LGLex* verbal and nominal entries have been converted into Alexina [TOL 11b]. Grammatical functions of arguments, not explicitly encoded in the LG tables or in *LGLex*, have been formalised for *LGLex*-Le*fff*. Information about prepositions, which can be expressed at three different levels in the LG (the lexical entry, the argument or the syntactic construction), has been copied for all arguments. The mnemonic identifiers of constructions have been parsed to deduce realisations of arguments. Both the *LGLex*-Le*fff* lexicon and the *LGLex*-to-Alexina converter are fully available[5] under a free license (LGPL-LR).

The LMF format is similar to Alexina. We implemented a similar *LGLex*-to-LMF converter for verbal entries.

---

[4] http://infolingu.univ-mlv.fr/english (Language Resources > Lexicon-Grammar > Download).
[5] The URL is the same as for the other LG resources.

### 11.3. Lexical Entries

The LG distinguishes lexical items on the basis of syntactic and semantic behaviour. For example, the verbs *voler* ‘fly’ and *voler* ‘steal’ are described in distinct lexical items. In this experiment of conversion, we generated LMF lexical entries in one-to-one correspondence with the LG items. Grouping them in function of their lemma and inflectional morphology would comply better with the LMF model, but we left this work for further versions.

Thus, the construction of the *LexicalEntry* elements from *LGLex* was mostly straightforward. Here is one of the resulting elements:

```
<LexicalEntry id="V_32RA_96" status="to be completed">
  <feat att="partOfSpeech" val="verb"/>
  <Lemma>
    <feat att="writtenForm" val="confirmer"/>
    <feat att="translation" val="to confirm"/>
    <feat att="example" val="Max a confirmé (la commande+le rendez-vous)"/>
  </Lemma>
  <SyntacticBehaviour
    subcategorizationFrameSets =
      "[Suj:cln|scompl|sinf|sn,Obj:sn|cla];@avoir,@ObjN-hum,@SujN-
hum,@SujNhum;%actif,%passif"/>
</LexicalEntry>
```

We generated the *id* attribute, which is the entry identifier, by concatenating the identifiers of its grammatical category and of the class it belongs to, and the number of the entry in the table. For instance, the *V_32RA_96* identifier corresponds to the 96th entry in verb class 32RA.

We added a *status* attribute which depends on the proportion of encoded features. The values are *"completed"* for a fully encoded entry, *"to be completed"* for an entry with at least one feature unencoded,[6] or *"to be encoded"* for an entry with less than 1/3 of encoded features.

We also added in *Lemma* an *example* which contains a typical sentence illustrating the verb use described in the entry, e.g. *Max a confirmé (la commande + le rendez-vous)* ‘Max confirmed the (order + meeting)’. When available, the *translation* contains an English gloss of the lemma.

The *SyntacticBehaviour* element allows for pointing to syntactic constructions either individually, through the *subcategorizationFrame* element, or by groups, through the *subcategorizationFrameSet* element. We decided to use only the latter possibility, and to group constructions as in *LGLex*-Le*fff*, i.e. only when they are closely related, e.g. an active construction and the corresponding passive construction. Thus, the *subcategorizationFrameSets* attribute contains space-separated identifiers of groups of constructions. The following *SyntacticBehaviour* element, extracted from the entry of *se hâter* ‘hasten’, identifies two groups, one for this verb’s constructions with nominal complements, as in *Max se hâte dans son travail* ‘Max hurries up in his work’, and the other for those with infinitival complements, as in *Max se hâte de répondre* ‘Max hastens to answer’:

```
<SyntacticBehaviour
  subcategorizationFrameSets =
    "[Suj:cln|sn,Obl:dans-sn];@pron,@être,@SujNhum;%actif
    [Suj:cln|sn,Obl:(de-sinf)];@pron,@être,@SujNhum,@CtrlSujObl;%actif"/>
```

[6] An unencoded feature is a feature present in the table, but assigned the ‘~’ code for the entry. The *status* attribute ignores the ‘O’ symbol in the table of classes, which also means that the feature should be encoded, but is not yet listed in the table.

Some French verbal items are lexically frozen with a non-argumental clitic pronoun, as *en coûter* ‘be costly’ in

(1) *De tels gestes en coûtent à leur auteur* ‘Such acts are costly to their author’

In this expression, the clitic pronoun *en* ‘of it’ does not refer to any entity, nor commute with a prepositional phrase. It is frozen with the verb. The LG represents such items in classes of simple verbs, by analogy with inherently pronominal verbs, as in

(2) *De tels gestes se retournent contre nous* ‘Such acts turn against us’

and mandatorily negative verbs, as in

(3) *Max ne décolère pas de cette erreur*
‘Max’s anger about this mistake does not abate’

However, frozen clitic/verb sequences as *en coûter* in (1) are multiword expressions (MWEs). Thus, we opted for encoding them with the LMF package for MWE patterns:

```
<LexicalEntry id="V_5_25" status="to be completed" mwePattern="en-V_y-V">
  <feat att="partOfSpeech" val="verb"/>
  <Lemma>
    <feat att="writtenForm" val="coûter"/>
    <feat att="example" val="Faire ce genre de truc en coûte à Luc"/>
  </Lemma>
  <ListOfComponents>
    <Component entry="PRO_en"/>
    <Component entry="V_coûter"/>
  </ListOfComponents>
  <SyntacticBehaviour
    subcategorizationFrameSets="[Suj:cln|scompl|sinf|sn,Obl:(à-sn|sn)];@avoir,@SujN-
hum,@OblNhum;%actif,%actif_impersonnel"/>
</LexicalEntry>
```

The 96 expressions listed in the LG required 4 MWE patterns such as the following:

```
<MWEPattern id="en-V_y-V">
  <MWENode>
    <MWEEdge>
      <feat attr="function" val="adjunct"/>
      <MWENode>
        <feat attr="syntacticConstituent" val="clitic-pronoun"/>
        <MWELex>
          <feat attr="componentRank" val="1"/>
        </MWELex>
      </MWENode>
    </MWEEdge>
    <MWELex>
      <feat attr="componentRank" val="2"/>
    </MWELex>
  </MWENode>
</MWEPattern>
```

## 11.4. Subcategorization frames

### 11.4.1. *Subcategorization frame sets*

Lexical entries point to subcategorization frame sets through identifiers. The *SubcategorizationFrameSet* class has an attribute which lists space-separated identifiers of *subcategorizationFrame* elements:

```
<SubcategorizationFrameSet
  id="[Suj:cln|sn,Obj:sn];@être,@ObjN-hum,@SujNhum;%actif,%passif"
  subcategorizationFrames=
    "[Suj:cln|sn,Obj:sn];@être,@ObjN-hum,@SujNhum;%actif
     [Suj:cln|sn,Obj:sn];@être,@ObjN-hum,@SujNhum;%passif"/>
```

Each *SubcategorizationFrame* is described in an XML element which bears the corresponding identifier, here the construction of verbs like *pouvoir* ‘can’:

```
<SubcategorizationFrame
  id="[Suj:cln|sn,Obl:sinf];@avoir,@SujN-hum,@SujNhum,@CtrlSujObl;%actif">
  <LexemeProperty>
    <feat att="voice" val="active"/>
    <feat att="auxiliary" val="avoir"/>
  </LexemeProperty>
  <SyntacticArgument>
    <feat att="id" val="0"/>
    <feat att="syntacticFunction" val="subject"/>
    <feat att="syntacticConstituent" val="clitic-nominative NP"/>
    <feat att="restriction" val="human non-human"/>
  </SyntacticArgument>
  <SyntacticArgument>
    <feat att="id" val="1"/>
    <feat att="syntacticFunction" val="object"/>
    <feat att="syntacticConstituent" val="infinitive-clause"/>
    <feat att="control" val="0"/>
  </SyntacticArgument>
</SubcategorizationFrame>
```

The *LexemeProperty* element provides four types of information.

The *auxiliary* indicates the auxiliary verbs for compound tenses: *avoir* or *être.* The verb *achever* takes *avoir*: *Max a achevé de peindre le mur* ‘Max has finished painting the wall’. The verb *s'arrêter* takes *être*: *Max s'est arrêté de boire* ‘Max stopped drinking’.

The *voice* specifies the morphological voice of the verb in the construction: *active* or *passive*.

The *negation* marks obligatorily negative verbs as in (3) (see Section 11.3).

The *non-argumental-clitic* specifies a clitic pronoun present in the construction: it takes the values *reflexive*, for pronominal constructions such as (2), and *impersonal*, for *il*-constructions as in (6) (see Section 11.4.3).

### 11.4.2. *Grammatical functions*

The LG, including *LGLex*, do not use the full set of grammatical functions taught by traditional grammar, but only subject, object and (implicitly) adjunct.[7] This option is motivated by the fact that, beyond these three functions, the remaining information conveyed by grammatical functions is redundant with other indispensable elements of description. For example, the distinction between direct object and indirect object is encoded in parallel in syntactic constituents, respectively specified as *NP* or *PP*. Distinctions between various types of non-prepositional objects are redundant with information about passive constructions, clitic pronominalization etc., which is encoded with more detail, in the LG, in the form of transformational features [GRO 69].

[7] Adjuncts are not represented in lexical entries, since they are, in general, little dependent on the lexical value of the predicate.

Thus, we retained a minimal set of grammatical functions. In order to comply with the data category register, we opted for: subject, object, agent, and inverted subject. The deduction of grammatical functions was adapted from the *LGLex*-to-Alexina converter.

### 11.4.3. *Representation of syntactic arguments*

Another salient difference between the LG model and the LMF format is the representation of syntactic arguments. In the tables and in *LGLex*, arguments are represented at the level of lexical entries, independently of the grammatical functions that they assume in specific constructions. Take, for example, the lexical item of *arriver* 'happen' exemplified by the sentence *De tels évènements arrivent souvent à Max* 'Such events often happen to Max'. The abstract argument, here *de tels évènements*, is described with the aid of distributional features which specify that it can be filled by non-human nouns, *que*-complementized completive clauses and infinitive clauses, but not by human nouns. These features are encoded by mnemonic identifiers such as $N_0$ =: *N-hum*, $N_0$ =: *Que P* etc. In parallel, constructions are described for the lexical item by independent features:

(4) *De tels évènements arrivent souvent à Max* 'Such events often happen to Max'
(5) *De tels évènements arrivent souvent* 'Such events often happen'
(6) *Il arrive souvent de tels évènements à Max* 'Such events often happen to Max'

As a matter of fact, the distributional features remain unchanged when this argument shifts to the position of inverted subject,[8] as in (6). They are represented by the same features for *arriver* as for verbal items which do not enter in construction (5), like *incomber* 'be the responsibility of'.

In LMF and Alexina, distributional features can only occur at the level of syntactic constructions. Thus, our converter duplicates them, which introduces redundancy in the dictionary. The same holds for other argument-specific features, such as the value of the preposition that introduces the human argument of (1) (see Section 11.3), *à* 'to': the feature remains unchanged in (6), but must be duplicated. This solution is compatible with current syntactic parsers, but, in addition to redundancy, it poses two technical problems.

a) How to track an argument across constructions? For example, how to encode formally that the subject of (4) is the same syntactic argument as the inverted subject of (6)? In LMF, the positions of a given argument in constructions can be mapped through *synArgMaps* elements. However, each *synArgMaps* element is valid only for one argument in two constructions, which makes this device complex to handle in practice. Tracking 2 arguments across 4 constructions, for example, would have required up to 12 *synArgMaps* elements. We decided not to use it for this first experiment of LMF conversion.

[8] In general, the nominal distribution of an argument is not altered when we switch from a construction to another in the same item. This fundamental fact led Zellig Harris to define his notion of transformation, and thus was the origin of transformational theories of syntax.

b) How to refer to an argument? The typical situation involved is the description of control, i.e. co-reference with the implicit subject of infinitive clauses. For example, if the subject of (4) is an infinitive clause, the implicit subject of the infinitive clause is interpreted as being the other argument of the main verb:

(7) *Bégayer arrive souvent à Max* ‘Stuttering often happens to Max’

In order to describe this as a feature of one of the two arguments, we need to refer to the other. LMF does not normalize a way of referring to a syntactic argument. Alexina does this through the grammatical function of the target argument, for example direct object and indirect object. The set of grammatical functions described in Section 11.4.2 is too reduced for this purpose. In addition, even with the full traditional set of grammatical functions, this solution does not work in all cases: when a verb has two arguments with the same grammatical function, e.g. two prepositional objects, this method of identification confounds them. In such cases, *LGLex*-Le*fff* resorts to additional grammatical functions, such as Obl2 and Obl3, for second or third indirect object, but the assignment of such functions is arbitrary. Thus, we innovated. We systematically numbered arguments in syntactic constructions, beginning from 0, through a *feat* element with an *id* attribute. In the argument containing the infinitive clause, we inserted a *feat* element with a *control* attribute, containing the number of the argument that refers to the implicit subject of the infinitive clause. For example, the construction of (7) is encoded as:

```
<SubcategorizationFrame
  id="[Suj:cln|scompl|sinf|sn,Obj:(à-sn|sn|cla)];@être,@SujN-hum,@ObjNhum;%actif">
  <LexemeProperty>
    <feat att="voice" val="active"/>
    <feat att="auxiliary" val="être"/>
  </LexemeProperty>
  <SyntacticArgument>
    <feat att="id" val="0"/>
    <feat att="syntacticFunction" val="subject"/>
    <feat att="syntacticConstituent"
      val="clitic-nominative completive-clause infinitive-clause NP"/>
    <feat att="restriction" val="non-human"/>
    <feat att="control" val="1"/>
  </SyntacticArgument>
  <SyntacticArgument>
    <feat att="id" val="1"/>
    <feat att="syntacticFunction" val="object"/>
```

```
    <feat att="syntacticConstituent" val="PP NP clitic-accusative"/>
    <feat att="optionality" val="optional"/>
    <feat att="restriction" val="human"/>
  </SyntacticArgument>
</SubcategorizationFrame>
```

When two arguments may control the infinitive clause, as in the following sentences:

*Tu éreintes les enfants à les faire courir partout*
'You are exhausting the kids by having them run everywhere'
*Tu éreintes les enfants à se préparer leur abri*
'You are exhausting the kids with preparing their shelter'

the numbers of the possible controllers are listed in the *feat* element with a *control* attribute, as in: val="0 1".

#### 11.4.4. *Levels of generality of syntactic constructions*

In the LG, constructions can be shared between entries even if details differ, through underspecification. For example, the feature that specifies the following syntactic construction:

(6) *Il arrive souvent de tels évènements à Max* 'Such events often happen to Max'

is also used for verbs that, in contrast with *arriver*, have no object, or accept a subject denoting a human:

(8) *Il éclata un orage* 'A storm came up', lit. 'It came up a storm'
(9) *Il a candidaté à ce poste vingt personnes*
'Twenty people candidated for this position',
lit. 'It candidated twenty people for this position'

The feature specifies only that the original subject shifts to another non-prepositional position, and an impersonal subject is inserted. The presence of an object and the distribution of the subject position are specified by independent features. Such underspecified features avoid prejudicial redundancy and contribute to make the LG readable: each of them is compact, and a set of less than 500 features is enough to encode all the information provided on verbs.

This style of encoding might be implemented in the LMF format, thanks to the possibility of inheritance between *subcategorizationFrame* elements. However, we left this perspective for future experiments. Most available LMF-encoded examples encode a given construction into a single *subcategorizationFrame* element, representing it at the maximal available level of detail. Developing a use of inheritance and the corresponding converter would have been a more innovative project, and required more time.

Thus, we encoded fully specified syntactic constructions, copying argument-specific features into construction-specific features. This led us to generate as many as 4 700 distinct constructions, for 13 900 lexical items (34%). In order to help human readers to manage such a bulk of data, we adopted mnemonic identifiers instead of numbers: namely, a variant of the Alexina encodings of the constructions,[9] by running the *LGLex*-to-Alexina converter in parallel with the *LGLex*-to-LMF converter. Each identifier contains: the list of arguments with their realisations; diverse feature labels; and labels for argument redistributions such as active or passive:

[Suj:cln|sn,Obl:(de-sinf)];@pron,@être,@SujNhum,@CtrlSujObl;%actif

#### 11.4.5. *Constituents*

The *syntacticConstituent* specifies the syntactic category of the constituent: noun phrase (*NP*), prepositional phrase (*PP*), *infinitive-clause, que*- or *le fait que*-complementized argument clause (*completive-clause*), *si*-complementized argument clause (*wh-completive-clause*), adjectival phrase (*adj*) and various types of clitic pronouns.

The *introducer* lists prepositions and specifies the possibility of locative prepositions such as *dans* ‘in’, *sur* ‘on’, *sous* ‘under’, *vers* ‘to’ etc.

The *restriction* specifies human or non-human semantic features of noun phrases and prepositional phrases. With most verbs, some animals are linguistically assimilated to persons [GUI 86].

We added an *optionality*, a *mood* for argument clauses: *indicative* or *subjunctive*, a *control* (cf. 11.4.3), and a *role* attribute which is filled for locative arguments realised as prepositional phrases.

### 11.5. Results

The LG of French verbs contains 13 900 lexical items, which describe 5 740 morphologically different verbs. Our conversion to LMF is automated, so that new versions can be generated when tables are updated. The LMF converter produces an 11-MB XML document, LG-LMF, with 4 700 *subcategorizationFrame* elements, grouped in 2 800 *subcategorizationFrameSet* elements (880 of them with one construction, 1 700 with two, 210 with three and 1 with four). The group with 4 constructions is for the verb *pardonner* ‘forgive’. LG-LMF is fully available under a free license (LGPL-LR) at http://infolingu.univ-mlv.fr/english (Language Resources > Lexicon-Grammar > Download).

Due to time limitations, some information provided in the LG was lost in this first experiment of conversion.

---

[9] We essentially substituted “[” for “<” and “]” for “>”.

Among distributional information consisting of semantic features, we retained only human and non-human noun phrases. More information is not useful to syntactic parsers, since current dictionaries lack a semantic classification of nouns.

We simplified the information about prepositions introducing completive clauses and infinitive clauses. In some French verbs, prepositional complements filled by completive clause can take a non-prepositional form:

*Max doute de la présence du chef* ‘Max doubts the presence of the boss’
*Max doute que le chef soit présent* ‘Max doubts the boss is present’

and direct complements filled by infinitive clauses can take a prepositional form:

*Max prévoit qu’il reviendra* ‘Max foresees he will come back’
*Max prévoit de revenir* ‘Max foresees to come back’

The transposition from the LG model to the LMF format is complex and we simplified it in order to avoid multiplying syntactic constructions.

We also dropped the complex controls not covered by our numbering of arguments, for example the control of infinitive clauses by a prepositional modifier of an argument:

*Max étend mes attributions à recevoir les paiements*
‘Max extends my duties to receiving payments’

This work was also an opportunity to detect errors in the LG. Some inherently pronominal verbs were encoded as having a passive construction, or as combining with the auxiliary verb *avoir* ‘have’ for compound tenses: this was corrected. A new syntactic feature, $N_0$ *V de* $N_2$, was substituted for $N_0$ *V Prép* $N_2$ in class 13, since the value of the preposition could not be retrieved from other features.

## 11.6. Conclusion

We described the conversion of the Lexicon-Grammar (LG) of French verbs into the LMF format. This work contributes to the standardization of lexical resources and their interoperability at the lexical-syntactic level for French. All conversion tools, and the LMF version of the LG,referred to as LG-LMF, are fully available under a free license (LGPL-LR) at http://infolingu.univ-mlv.fr/english (Language Resources > Lexicon-Grammar > Download).

This work was also an opportunity for us to compare the LG model with the LMF format. They have distinct objectives, and they differ in the way of managing redundancy. The LMF representation of syntax is based on the notion of syntactic construction. Most syntactic information must be attached to syntactic constructions, and this implies duplicating it. The resulting data are less readable than formats dedicated to maintenance or creation of dictionaries by linguists, such as that of LG tables, which are structured on the notion of syntactic feature. A solution is to perform updates on a dictionary with high readability, like the LG tables, and to compile it after each operation, in the same manner as a dictionary of lemmas is updated and compiled into a dictionary of inflected forms.

# Bibliography

[BOO 76] Boons J.-P., Guillet A., Leclère C., *La structure des phrases simples en français : Constructions intransitives*. Droz, Genève, Suisse, 1976.

[CON 10] Constant M., Tolone E., A generic tool to generate a lexicon for NLP from Lexicon-Grammar tables, in Gioia M. D. (ed.), *Actes du 27e Colloque international sur le lexique et la grammaire (L'Aquila, 10-13 septembre 2008), Seconde partie*, volume 1 de Lingue d'Europa e del Mediterraneo, Grammatica comparata, pages 79–193, Aracne, Rome, Italie, 2010.

[GRO 69] Gross M., Remarques sur la notion d'objet direct en français. *Langue Française* 1, pp. 63-73, Paris: Larousse, 1969.

[GRO 75] Gross M., *Méthodes en syntaxe : Régimes des constructions complétives*. Hermann, Paris, France, 1975.

[GUI 86] Guillet, A., Représentation des distributions dans un lexique-grammaire. *Langue Française* 69, Paris: Larousse, 1986.

[GUI 92] Guillet A., Leclère C., *La structure des phrases simples en français : Les constructions transitives locatives*. Droz, Genève, Suisse, 1992.

[HAR 57] Harris, Z.S., Co-occurrence and transformations in linguistic structure. *Language* 33, pp. 283-340, 1957.

[HAT 98] Hathout, N., Namer, F., Automatic construction and validation of French large lexical resources: reuse of verb theoretical linguistic descriptions, in: *Proceedings of the Language Resources and Evaluation conference* (Granada, Spain, May 1998), European Language Resource Association (ELRA), Paris, pp. 627-636, 1998.

[LEC 02] Leclère C., Organization of the Lexicon-Grammar of French verbs, *Lingvisticae Investigationes* 25:1, pages 29–48, Amsterdam/Philadelphia : John Benjamins, 2002.

[SAG 10] Sagot B., The Le*fff*, a freely available and large-coverage morphological and syntactic lexicon for French, in: *Proceedings of the Language Resources and Evaluation Conference* (Valletta, Malta, May 2010), European Language Resource Association (ELRA), Paris, 2010.

[TOL 11a] Tolone E., *Analyse syntaxique à l'aide des tables du Lexique-Grammaire*, PhD thesis, 340 pp., LIGM, Université Paris-Est, France, 2011.

[TOL 11b] Tolone E., Sagot B., Using Lexicon-Grammar tables for French verbs in a large-coverage parser, in Vetulani Z. (ed.), *Human Language Technology, Challenges for Computer Science and Linguistics, 4th Language and Technology Conference, LTC 2009,* Poznań, Poland, November 2009, Revised Selected Papers, Lecture Notes in Artificial Intelligence (LNAI), Springer Verlag, vol. 6562, pp. 183-191, 2011.